\documentclass[letterpaper]{article} 
\usepackage{aaai2026}
\usepackage{times}  
\usepackage{helvet}  
\usepackage{courier}  
\usepackage[hyphens]{url}  
\usepackage{graphicx} 
\usepackage{natbib}  
\usepackage{caption} 
\usepackage{algorithm}
\usepackage{algorithmic}
\usepackage{booktabs} 
\usepackage{placeins}
\usepackage{amsmath}
\usepackage{amssymb}
\usepackage{newfloat}
\usepackage{listings}
\DeclareCaptionStyle{ruled}{labelfont=normalfont,labelsep=colon,strut=off} 
\floatstyle{ruled}
\newfloat{listing}{tb}{lst}{}
\floatname{listing}{Listing}
\title{PoeTone: A Framework for Constrained Generation of Structured Chinese Songci with LLMs}
\author{
    Zhan Qu,
    Shuzhou Yuan,
    Michael Färber
}
\affiliations{
    TU Dresden \\ ScaDS.AI \\

    Dresden, Saxony, Germany\\
    zhan.qu@tu-dresden.de
}

\usepackage{bibentry}

\begin{document}

\maketitle

\begin{abstract}
This paper presents a systematic investigation into the constrained generation capabilities of large language models (LLMs) in producing \textit{Songci}, a classical Chinese poetry form characterized by strict structural, tonal, and rhyme constraints defined by Cipai templates. We first develop a comprehensive, multi-faceted evaluation framework that includes: (i) a formal conformity score, (ii) automated quality assessment using LLMs, (iii) human evaluation, and (iv) classification-based probing tasks. Using this framework, we evaluate the generative performance of \textbf{18 LLMs}, including 3 proprietary models and 15 open-source models across 4 families, under \textbf{five prompting strategies}: zero-shot, one-shot, completion-based, instruction-based, and chain-of-thought. Finally, we propose a Generate-Critic architecture in which the evaluation framework functions as an automated critic. Leveraging the critic’s feedback as a scoring function for best-of-N selection, we fine-tune 3 lightweight open-source LLMs via supervised fine-tuning (SFT), resulting in improvements of up to \textbf{5.88\%} in formal conformity. Our findings offer new insights into the generative strengths and limitations of LLMs in producing culturally significant and formally constrained literary texts.
\end{abstract}

\begin{links}
    \link{Code}{https://github.com/ZhanQu945/PoeTone}
\end{links}

\section{Introduction}

Recent advancements in large language models (LLMs) have demonstrated remarkable capabilities in generating fluent, coherent, and contextually appropriate text across diverse domains \cite{brown2020language, touvron2023llama}. From drafting emails and summarizing documents to composing fiction and poetry, models such as \texttt{GPT-4o}, \texttt{Gemini 2.5 Pro}, and \texttt{DeepSeek-R1} have significantly expanded the creative potential of AI systems \cite{openai2024gpt4technicalreport, comanici2025gemini25pushingfrontier, deepseekai2025deepseekr1incentivizingreasoningcapability}. Yet, tasks requiring both expressive fluency and strict adherence to formal rules, such as generating classical poetry, remain a significant challenge \cite{yu2024charpoetchineseclassicalpoetry}.

\begin{figure}[ht!]
    \centering
    \includegraphics[width=0.85\linewidth]{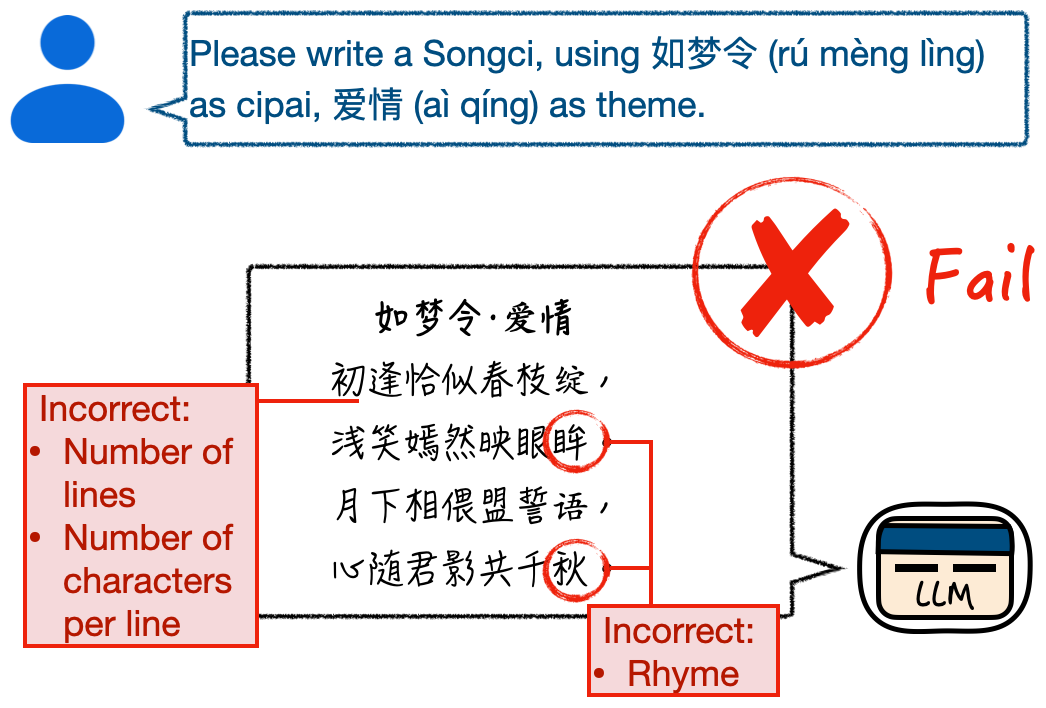}
    \caption{The multi-layered challenge of generating high-quality \textit{Songci}, highlighting the formal constraints of the Cipai as the primary bottleneck.}
    \label{fig:constraint_funnel}
\end{figure}

This paper focuses on \textit{Songci}, a prominent form of Chinese lyric poetry from the Song dynasty (960–1279 CE). Unlike Tang poetry, \textit{Songci} follows fixed \textit{Cipai}, predefined tune patterns that govern stanza count, line count, line length, tonal flow (píng and zè), rhyme type, and rhyme positions \cite{ge2009cilin}. Effective \textit{Songci} must not only conform to these structural constraints but also express coherent and emotionally resonant themes, typically involving longing, solitude, patriotism, or nature. This blend of metrical rigidity and artistic depth makes \textit{Songci} a compelling testbed for constrained text generation. A common failure case in generating \textit{Songci} is shown in Figure~\ref{fig:constraint_funnel}.

While prior work has applied deep learning to classical Chinese poetry, most focus on stylistic imitation or surface-level coherence \cite{LiaoGPTbasedpoetry, tang2025simple}. Whether modern LLMs can generate \textit{Songci} that satisfy its structural and tonal requirements remains largely unexplored \cite{song-2022-composing}. Moreover, existing benchmarks lack fine-grained tools for evaluating formal correctness, and few approaches aim to improve generation under these constraints.

\begin{figure*}[ht!]
    \centering
    \includegraphics[width=0.8\linewidth]{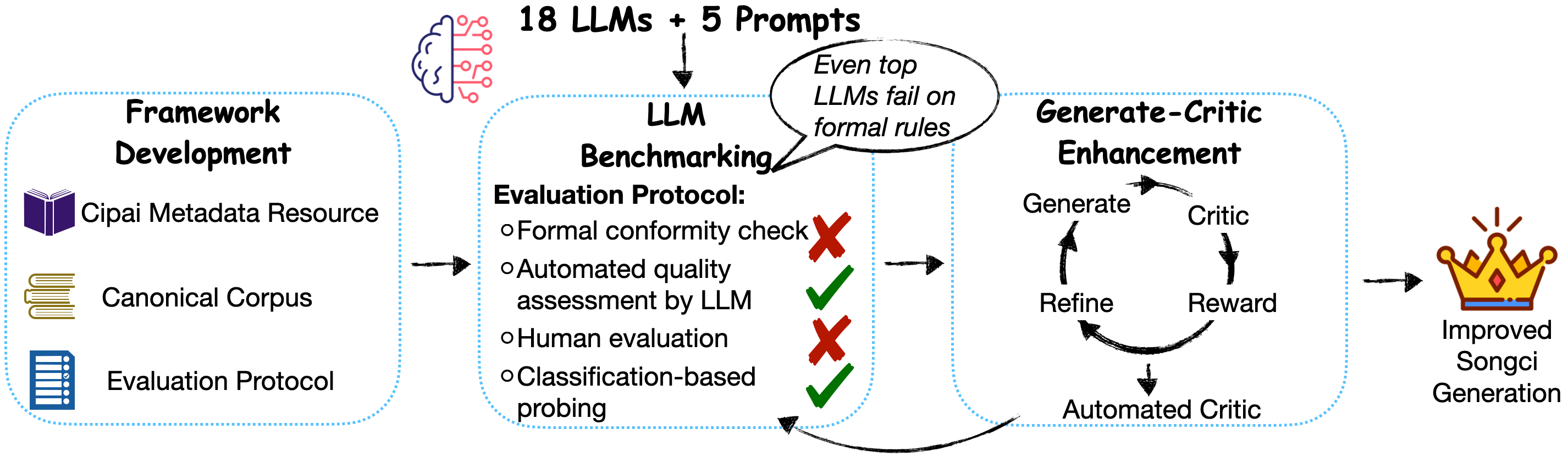}
    \caption{An overview of our research pipeline, from framework development and benchmarking to model enhancement.}
    \label{fig:research_pipeline}
\end{figure*}

To bridge these gaps, we introduce \textbf{PoeTone}, a systematic investigation into the capacity of LLMs to generate \textit{Songci} while adhering to its strict metrical constraints. Our complete research pipeline is detailed in Figure~\ref{fig:research_pipeline}:

First, we develop a comprehensive evaluation framework that includes: (i) a metadata resource detailing the structure and tonal patterns of 20 widely used \textit{Cipai}, (ii) a curated corpus of canonical \textit{Songci} poems organized by theme and \textit{Cipai}, and (iii) a multi-dimensional evaluation protocol combining formal conformity scores, automated quality assessment, human ratings, and classification-based probing.

Second, we benchmark 18 state-of-the-art LLMs, including 3 proprietary models (\texttt{GPT-4o}, \texttt{Gemini 2.5 Pro}, \texttt{ERNIE 4.5 Turbo}) and 15 open-source models from 4 families (\texttt{LLaMA}, \texttt{Mistral}, \texttt{Qwen}, \texttt{DeepSeek}), across five prompting strategies: zero-shot, one-shot, completion, instruction, and chain-of-thought.

Third, we propose a Generate-Critic architecture that uses rule-based conformity scoring to guide fine-tuning. Applied to 3 open-source LLMs via best-of-$N$ rejection sampling and LoRA-based supervised fine-tuning (SFT), this method improves adherence to structural and tonal constraints by up to \textit{5.88\%}.

\textbf{In summary, our key contributions are:}
\begin{itemize}
    \item A structured evaluation framework for \textit{Songci} generation, including a metadata resource, a canonical corpus, and a multi-faceted evaluation framework.
    \item A benchmark of 18 LLMs under five prompting strategies, revealing model limitations in conforming to formal poetic constraints.
    \item A Generate-Critic method that improves constrained generation through automated rule-based feedback.
\end{itemize}

\section{Related Work}
Automatic generation of Chinese poetry has evolved from statistical and rule-based methods to neural networks \cite{wang-etal-2016-chinese, chen2024polishing}. Initial approaches used RNNs and LSTMs \cite{zhang2014chinese, wang-etal-2016-chinese}, later enhanced by attention mechanisms to improve coherence and context modeling \cite{yi2017generating}.

A major research focus has been controlling poetic form and content. To handle strict structural and rhyming constraints, methods include dual-encoder models conditioned on rhythmic patterns \cite{luo2021chinese} and form-aware generation guided by stressed weighting \cite{hu2020generating}. Thematic coherence has been addressed through planning-based structures \cite{wang-etal-2016-chinese}, working memory models \cite{yi2018chinese}, salient clue guidance \cite{yi-etal-2018-chinese, gao2021new}, adversarial training for title consistency \cite{li2018generating}, and graph neural networks for topic modeling \cite{yan-etal-2023-poetry}. Sentiment and style control have been explored via multi-tag classification and conditional variational autoencoders \cite{shao2021sentiment, chen2019sentiment, MixSong}.

The rise of large language models (LLMs) has transformed Chinese poetry generation. An early study showed that a simple Generative Pre-trained Transformer (GPT) could produce high-quality classical poetry without extensive feature engineering \cite{LiaoGPTbasedpoetry}. In further studies, researchers have fine-tuned pre-trained models on specialized poetry corpora, finding that smaller, domain-specific models can sometimes outperform larger, general-purpose LLMs like GPT-4o on poetic tasks \cite{tang2025simple, wang2022generationchineseclassicalpoetry}. Innovations like CharPoet introduce token-free, character-level control to improve format accuracy \cite{yu2024charpoetchineseclassicalpoetry}, reflecting a trend to harness LLM capabilities while imposing classical poetic constraints.

Beyond text generation, interactive systems such as Jiuge \cite{zhipeng-etal-2019-jiuge} and Yu Sheng \cite{ma-etal-2023-yu} incorporate human-in-the-loop frameworks for collaborative poem refinement. SongSong \cite{hu2025songsong} generates music from \textit{Songci} lyrics, addressing musical restoration, but leaving high-quality, metrically-correct \textit{Songci} generation as an open challenge. Our work focuses on this core task, leveraging LLMs to generate formal \textit{Songci} text.

\section{Songci Generation and Evaluation Framework}
Evaluating the ability of LLMs to generate valid \textit{Songci} requires two key resources that, to our knowledge, are unavailable in structured, machine-readable form. First, a formal constraint specification for each \textit{Cipai} is necessary for automated verification. Second, a high-quality corpus of canonical \textit{Songci}, thematically categorized and representative of diverse \textit{Cipai}, is essential for model prompting and evaluation. This section details the creation of both resources and introduces our multi-faceted evaluation protocol.

\subsection{Cipai Formal Constraint Resource}
The primary challenge in \textit{Songci} generation lies in adhering to the strict structural and phonological rules defined by each \textit{Cipai}. To enable automated verification of these constraints, we constructed a structured metadata resource covering the 20 most commonly used two-stanza \textit{Cipai}, selected for their frequency in classical anthologies and representational diversity~\cite{ge2009cilin}. Each \textit{Cipai} may have up to 15 different variants, and all of them are included in the metadata. 

Each \textit{Cipai} is encoded in machine-readable JSON format, capturing the following elements:
\begin{itemize}
    \item \textbf{Structure}: Specifies the number of stanzas, total line count, and exact character count for each line (with segmentation based on punctuation, i.e., commas, periods, and question marks).
    \item \textbf{Tonal Pattern}: Encodes the expected tonal category, píng (level) or zè (oblique), at key positions per line, using modern Mandarin tonal approximations.
    \item \textbf{Rhyme Scheme}: Identifies lines and positions that must rhyme, with rhyme group constraints (píng or zè) derived from classical rhyme dictionaries such as \textit{Cilin Zhengyun}~\cite{ge2009cilin}.
\end{itemize}

\subsection{Thematic Canonical Corpus}
We compiled a curated corpus of 120 human-authored \textit{Songci} as exemplars for prompting and evaluation. The \textit{Songci} were selected from authoritative digital archives based on author prominence (e.g., Su Shi, Li Qingzhao) and fidelity to canonical \textit{Cipai} forms \cite{ge2009cilin}.

To support thematic generation and classification, each \textit{Songci} was manually annotated with one of six recurring themes in classical \textit{Songci}:
\begin{itemize}
    \item \textbf{Love \& Longing}: Romantic and delicate emotions.
    \item \textbf{Patriotism \& Heroism}: Courageous expressions.
    \item \textbf{Nature \& Landscapes}: Scenery, seasons, pastoral life.
    \item \textbf{History \& Nostalgia}: Reflection on past eras or figures.
    \item \textbf{Sorrow \& Grief}: Farewell, exile, homesickness.
    \item \textbf{Philosophical Reflection}: Tranquil and spiritual tones.
\end{itemize}

This dataset serves as reference and training data for some of our analyses.

\subsection{Multi-Faceted Evaluation Protocol}
\label{sec:eval_protocol}

Evaluating the quality of generated \textit{Songci} requires more than assessing fluency or content relevance. A \textit{Songci} that reads naturally may still violate strict structural, tonal, or rhyming constraints, making it invalid under its specified \textit{Cipai}. Conversely, a \textit{Songci} that rigidly conforms to format may fail to achieve poetic elegance or thematic depth. To address this duality, we design a layered evaluation framework comprising both objective rule-based metrics and subjective assessments of fluency and artistic merit.

\paragraph{Formal Conformity Score}
At the core of our evaluation is the \textbf{Formal Conformity Score}, which quantifies adherence to the structural, tonal, and rhyming rules encoded in the \textit{Cipai} metadata. The score is computed automatically using a rule-checking script that evaluates each generated \textit{Songci} along three dimensions:

\begin{itemize}
    \item \textbf{Structural Integrity}: It verifies that the \textit{Songci} has the correct number of lines and that each line has the precise number of characters required by the Cipai.
    \item \textbf{Tonal Adherence}: Using a standard modern Mandarin Chinese pronunciation lexicon, it checks if the characters at key positions match the mandatory píng or zè tonal patterns (or zhōng means both píng and zè are allowed).
    \item \textbf{Rhyme Scheme Compliance}: It identifies the characters at the designated rhyming positions and verifies that they belong to the same rhyme group according to a classical Chinese rhyme dictionary \cite{ge2009cilin}.
\end{itemize}

The quality of a generated \textit{Songci} is evaluated using a normalized, weighted scoring system. The final score is determined by calculating a `best-fit' against all known variants of a given \textit{Cipai} pattern. Formally, the total score $S_{\text{total}}$ for a generated \textit{Songci} is given by:

\begin{equation}
\label{eq:total_score}
S_{\text{total}} = \max_{v \in V} \left( w_S \cdot S_{\text{structure}}^{(v)} + w_T \cdot S_{\text{tonal}}^{(v)} + w_R \cdot S_{\text{rhyme}}^{(v)} \right)
\end{equation}

\noindent Where $V$ is the set of all known variants of the target \textit{Cipai}, $S_{\text{structure}}^{(v)}$, $S_{\text{tonal}}^{(v)}$, and $S_{\text{rhyme}}^{(v)}$ are component scores against variant $v$; $w_S$, $w_T$, and $w_R$ are non-negative weights summing to 1.0. We use $w_S = 0.4$, $w_T = 0.3$, and $w_R = 0.3$.

The \textbf{structure score}, $S_{\text{structure}}$, measures completeness and structural correctness of a generated \textit{Songci}. Let $N_g$ be the number of generated lines and $N_v$ be the number of template lines. The number of correctly structured lines, $N_{\text{match}}$, is found by checking the character count of a generated line, $|L_{g,i}|$, against the template's required count, $C_{v,i}$, for all lines up to the minimum of the two lengths.
\begin{equation}
N_{\text{match}} = \sum_{i=1}^{\min(N_g, N_v)} \mathbb{I}(|L_{g,i}| = C_{v,i})
\end{equation}
The final score is $N_{\text{match}}$ normalized by the maximum of the two lengths, which penalizes both missing and superfluous lines.
\begin{equation}
\label{eq:structure_score}
S_{\text{structure}} = \frac{N_{\text{match}}}{\max(N_g, N_v)}
\end{equation}

The \textbf{tonal score}, $S_{\text{tonal}}$, measures the quality of the well-formed parts of the \textit{Songci}. Let $I_{\text{match}}$ be the set of line indices that are structurally correct. Let the function $\text{tone}(c)$ return the tonal category of a character $c$, and let $T_{v,ij}$ be the required tone for the $j$-th character of the $i$-th line in the template. The equality $\text{tone}(c_{g,ij}) = T_{v,ij}$ is also considered true if $T_{v,ij}$ is marked as `zhōng' in the template. The score is the ratio of matching tones to the total characters in the evaluated lines.
\begin{equation}
\label{eq:tonal_score}
S_{\text{tonal}} = \frac{\sum_{i \in I_{\text{match}}} \sum_{j=1}^{|L_{g,i}|} \mathbb{I}(\text{tone}(c_{g,ij}) = T_{v,ij})}{\sum_{i \in I_{\text{match}}} |L_{g,i}|}
\end{equation}

The \textbf{rhyme score}, $S_{\text{rhyme}}$, quantifies the internal consistency of rhyme among the generated rhyming lines. Let $L_{g,R_v}$ denote the set of final characters from the generated \textit{Songci} corresponding to the rhyming positions specified by variant $v$ of the target \textit{Cipai}. Let $G$ be the set of rhyme groups defined in the metadata, and let $\text{rhyme}(c)$ return the group of character $c$. The score is defined as the proportion of rhyming characters that belong to the largest single rhyme group:
\begin{equation}
\label{eq:rhyme_score}
S_{\text{rhyme}} = \frac{\max_{g \in G} |\{ c \in L_{g,R_v} \mid \text{rhyme}(c) = g \}|}{|L_{g,R_v}|}
\end{equation}

\paragraph{Automated Quality Assessment}
To complement rule-based formal evaluation, we introduce an LLM-based assessment protocol to estimate the semantic and aesthetic quality of generated \textit{Songci} at scale. We use two powerful proprietary models (\texttt{GPT-4o} and \texttt{ERNIE 4.5 Turbo}), and calculate the average score to minimize bias. Each generated \textit{Songci} is presented to the judge model with its associated \textit{Cipai} and theme, and scored along three dimensions:

\begin{itemize}
    \item \textbf{Fluency}: Grammatical correctness and natural phrasing.
    \item \textbf{Coherence}: Logical and thematic consistency.
    \item \textbf{Poetic Quality (Yì Jìng)}: The overall artistic effect, emotional resonance, and imagistic depth.
\end{itemize}

Each criterion is rated on a 1–5 scale. This method provides a fast, repeatable, and cost-effective proxy for human evaluation. While the model-based judge may not fully replicate human literary sensibility, it enables consistent comparisons across thousands of outputs, and serves as a complementary lens to human judgment.

\paragraph{Human Evaluation and Poetic Turing Test}
To evaluate both the literary quality and perceived human-likeness of generated \textit{Songci}, we design a two-stage human evaluation study. We select a subset of generated \textit{Songci} from each LLM, chosen based on their highest combined Formal Conformity and LLM-judge scores. For each selected sample, we retrieve a matched human-written \textit{Songci} from the canonical corpus with the same \textit{Cipai} and theme, resulting in paired comparisons per model.

Each evaluation trial followed a two-stage procedure:

\textbf{(1) Poetic Turing Test:} Evaluators were shown a pair of anonymized \textit{Songci} (one generated, one human-written) and asked to identify which \textit{Songci} they believed was written by a human, followed by a 1–5 confidence score. 

\textbf{(2) Qualitative Scoring:} After the true authorship was revealed, evaluators rated each LLM-generated \textit{Songci} individually across three dimensions on a 1–5 Likert scale:
\begin{itemize}
    \item \textbf{Thematic Faithfulness}: Alignment with the assigned theme in mood and content.
    \item \textbf{Artistic Merit}: Aesthetic impact, use of imagery, and literary creativity.
    \item \textbf{Overall Quality}: A holistic assessment of form, expression, and emotional depth.
\end{itemize}

\paragraph{Classification-based Probing}
To investigate whether LLMs implicitly encode stylistic and thematic features in their outputs, we conduct three probing tasks using classifiers trained on our canonical corpus of 120 human-written \textit{Songci}: \textbf{Cipai Identification}, \textbf{Theme Classification}, and \textbf{Source Attribution (Human vs. LLM)}. 

We experiment with two lightweight classification pipelines: (1) a Support Vector Machine (SVM) trained on character-level embeddings from a pretrained Chinese BERT model, and (2) a Multinomial Naive Bayes classifier using unigram TF-IDF features.

\section{Benchmarking LLMs for \textit{Songci} Generation}
We designed a large-scale benchmark to evaluate the ability of contemporary LLMs to generate \textit{Songci} under formal and thematic constraints. Our evaluation spans a broad range of models and prompting strategies, enabling analysis of both intrinsic capabilities and responsiveness to guidance.

\subsection{Models Under Evaluation}
To facilitate comparison across model sizes, training regimes, and linguistic specifications, we selected 18 large language models for evaluation, covering both proprietary and open-source LLMs from multiple architectural families. We include 3 \textbf{proprietary models}:
\begin{itemize}
    \item \textbf{GPT-4o} (OpenAI): A leading multimodal model with strong general-purpose reasoning and instruction-following abilities \cite{openai2024gpt4technicalreport}.
    \item \textbf{Gemini 2.5 Pro} (Google): A flagship model with multilingual support and a long context window for structured prompts \cite{comanici2025gemini25pushingfrontier}.
    \item \textbf{ERNIE 4.5 Turbo} (Baidu): A Chinese-optimized model enhanced with external knowledge graphs for improved cultural grounding \cite{ernie2025technicalreport}.
\end{itemize}

We include 15 \textbf{open-source LLMs} from 4 major families:
\begin{itemize}
    \item \textbf{LLaMA (Meta)}: Strong general-purpose baselines, though English-centric tokenization may limit performance on Chinese text \cite{grattafiori2024llama}.
    \item \textbf{Mistral/Mixtral (Mistral AI)}: Efficient sparse mixture-of-experts (MoE) model (Mixtral) with high performance per parameter, but similarly limited in Chinese-specific pretraining \cite{jiang2024mixtralexperts}.
    \item \textbf{Qwen (Alibaba)}: Multilingual models with character-aware tokenization and strong performance on Chinese tasks \cite{yang2025qwen3technicalreport}.
    \item \textbf{DeepSeek (DeepSeek-AI)}: Chinese-based models optimized for reasoning and Chinese language understanding \cite{deepseekai2025deepseekr1incentivizingreasoningcapability}.
\end{itemize}

\subsection{Prompting Strategies} 
We designed five prompting strategies to assess how different forms of instruction impact generation quality. For each of the 20 \textit{Cipai} and 6 themes, every model generated \textit{Songci} using all five different prompts:

\begin{itemize}
    \item \textbf{Zero-shot}: Minimal instruction testing the model's latent knowledge of \textit{Songci}.
    \item \textbf{One-shot}: Provides a single example from our curated corpus to guide generation.
    \item \textbf{Completion}: The prompt includes the first stanza of a canonical \textit{Songci} to establish rhythm and style. The model completes the poem by generating the second stanza. Evaluation focuses solely on the generated half, and we ensure that it is not a direct reproduction of the original second stanza.
    \item \textbf{Instruction}: Presents explicit structural, tonal, and rhyme rules for the target \textit{Cipai}.
    \item \textbf{Chain-of-Thought (CoT)}: The model is prompted to first articulate the formal rules before generating the 
    \textit{Songci}, allowing us to test its capacity for self-directed reasoning. The reasoning text of the output is evaluated separately from the \textit{Songci} itself.
\end{itemize}


\begin{table*}[ht!]
\centering
\scalebox{0.8}{
\begin{tabular}{@{}l cccccc c@{}}
\toprule
\textbf{Model} & \textbf{Zero-shot} & \textbf{One-shot} & \textbf{Completion} & \textbf{Instruction} & \textbf{Chain-of-Thought} & \textbf{Average} & \textbf{Best Score} \\
\midrule
\multicolumn{8}{l}{\textit{Proprietary Models}} \\
\midrule
GPT-4o & \textbf{78.62} & 76.59 & 75.79 & 73.70 & 71.11 & 75.16 & 78.62 (zero-shot) \\
Gemini 2.5 Pro & 78.21 & 77.15 & 76.50 & 74.90 & 72.34 & 76.52 & 78.21 (zero-shot) \\
ERNIE 4.5 Turbo & 77.36 & \textbf{80.72} & \textbf{82.38} & \textbf{81.54} & \textbf{78.57} & \textbf{79.71} & \textbf{82.38 (completion)} \\
\midrule
\multicolumn{8}{l}{\textit{Open-Source Models}} \\
\midrule
DeepSeek-R1-Distill-Llama-70B & 33.75 & 41.71 & 38.10 & 35.95 & 35.58 & 37.82 & 41.71 (one-shot) \\
DeepSeek-R1-0528-Qwen3-8B & 51.34 & 49.09 & 51.18 & 42.50 & 27.61 & 44.74 & 51.34 (zero-shot) \\
Llama-3-8B-Instruct & 34.13 & 58.21 & 47.74 & 39.69 & 35.70 & 43.89 & 58.21 (one-shot) \\
Llama-3.1-8B-Instruct & 35.20 & 54.42 & 53.40 & 38.72 & 40.13 & 44.77 & 54.42 (one-shot) \\
Llama-3.2-1B-Instruct & 33.46 & 35.30 & 34.04 & 35.32 & 33.46 & 34.32 & 35.32 (instruction) \\
Llama-3.2-3B-Instruct & 36.15 & 46.02 & 45.79 & 35.35 & 36.47 & 39.96 & 46.02 (one-shot) \\
Llama-3.3-70B-Instruct & 35.88 & 63.07 & 51.16 & 37.12 & 35.31 & 44.91 & 63.07 (one-shot) \\
Mixtral-8x7B-Instruct-v0.1 & 28.04 & 41.10 & 35.63 & 30.34 & 31.11 & 33.24 & 41.10 (one-shot) \\
Mistral-7B-Instruct-v0.3 & 33.60 & 43.13 & 39.84 & 31.87 & 35.09 & 36.71 & 43.13 (one-shot) \\
Mistral-Small-Instruct-2409 & 37.11 & 54.41 & 53.42 & 40.42 & 35.26 & 44.12 & 54.41 (one-shot) \\
Qwen3-0.6B & 38.76 & 49.74 & 54.11 & 40.52 & 39.77 & 44.58 & 54.11 (completion) \\
Qwen3-1.7B & 39.88 & 48.56 & 47.20 & 44.20 & 40.72 & 44.51 & 48.56 (one-shot) \\
Qwen3-4B & 53.61 & 58.85 & 49.54 & 46.45 & 50.44 & 51.78 & 58.85 (one-shot) \\
Qwen3-8B & 60.19 & 64.53 & \textbf{63.75} & 53.98 & \textbf{62.58} & 60.61 & 64.53 (one-shot) \\
Qwen3-32B & \textbf{64.93} & \textbf{68.55} & 61.52 & \textbf{56.51} & 62.03 & \textbf{62.71} & \textbf{68.55 (one-shot)} \\
\bottomrule
\end{tabular}}
\caption{Formal Conformity Scores (\%) Across All Models and Prompting Strategies.}
\label{table:conformity}
\end{table*}

\subsection{Benchmarking Results}

\paragraph{Formal Conformity}
Table~\ref{table:conformity} presents the Formal Conformity Scores across all models and prompting strategies. Overall, proprietary models outperform open-source counterparts by a large margin. The best-performing proprietary model, \texttt{ERNIE 4.5 Turbo}, achieves the highest score of 82.38\% under the Completion prompt, followed by \texttt{Gemini 2.5 Pro} and \texttt{GPT-4o}, both of which show strong performance under Zero-shot prompting. Interestingly, \texttt{GPT-4o} performs best when given minimal guidance, suggesting its strong internal prior knowledge for classical poetic form.

Among open-source models, the \texttt{Qwen3} models are the top performers, with \texttt{Qwen3-32B} as the strongest, achieving a best score of 68.55\% under the One-shot prompt. Other open-source families (DeepSeek, LLaMA, Mistral) generally show lower average scores, indicating weaker internalization of \textit{Songci}’s structural constraints, likely due to tokenizer mismatch and lack of domain-specific pretraining.

Prompting strategies also influence performance significantly. One-shot prompting yields the highest average scores across open-source models, while Completion and Instruction perform best for \texttt{ERNIE}. Chain-of-Thought prompts were generally less effective. Manual inspection of the generated reasoning reveals that models often correctly identify the formal requirements of each \textit{Cipai}; however, the increased generation length may dilute adherence to structural constraints in the actual \textit{Songci}.

\begin{table*}[ht!]
\scalebox{0.75}{
\centering
\begin{tabular}{@{}l c ccc cccc@{}}
\toprule
 & & \multicolumn{3}{c}{\textbf{LLM-as-Judge Scores}} & \multicolumn{4}{c}{\textbf{Human Evaluation Scores}} \\
\cmidrule(lr){3-5} \cmidrule(lr){6-9}
\textbf{Model} & \textbf{Best Prompt} & \textbf{Fluency} & \textbf{Coherence} & \textbf{Poetic Quality} & \textbf{Turing Test} & \textbf{Thematic Faith.} & \textbf{Artistic Merit} & \textbf{Overall Quality} \\
\midrule
\multicolumn{9}{l}{\textit{Proprietary Models}} \\
\midrule
GPT-4o & zero-shot & 4.72 & \textbf{4.26} & \textbf{4.35} & 2.10 & \textbf{3.60} & \textbf{3.20} & 3.50 \\
ERNIE 4.5 Turbo & completion & \textbf{4.75} & 4.05 & 4.10 & \textbf{2.15} & 3.45 & 3.10 & \textbf{3.53} \\
\midrule
\multicolumn{9}{l}{\textit{Open-Source Models}} \\
\midrule
DeepSeek-R1-0528-Qwen3-8B & zero-shot & 4.10 & 3.80 & 3.60 & 1.80 & 2.90 & 2.70 & 2.85 \\
Llama-3.3-70B-Instruct & one-shot & 4.05 & 3.60 & 3.40 & 1.75 & 2.80 & 2.55 & 2.70 \\
Mistral-Small-Instruct-2409 & one-shot & 3.90 & 3.30 & 3.10 & 1.60 & 2.50 & 2.30 & 2.45 \\
Qwen3-32B & one-shot & 4.00 & 3.40 & 3.20 & 1.65 & 2.60 & 2.35 & 2.50 \\
\bottomrule
\end{tabular}}
\caption{Poetic Quality Scores (1–5 scale) from LLM-as-Judge and Human Evaluation.}
\label{table:quality}
\end{table*}

\paragraph{Poetic Quality and Human Judgement}
To assess the stylistic expressiveness and thematic alignment of generated \textit{Songci} beyond structural correctness, we selected a representative subset of models for deeper evaluation. From the proprietary group, we chose \texttt{GPT-4o} and \texttt{ERNIE 4.5 Turbo}, both of which demonstrated the highest average Formal Conformity Scores. From the open-source group, we selected one top model from each major family: \texttt{DeepSeek-R1-0528-Qwen3-8B}, \texttt{Llama-3.3-70B-Instruct}, \texttt{Mistral-Small-Instruct-2409}, and \texttt{Qwen3-32B}. For each model, we used its best-performing prompt as determined by the conformity evaluation.

Automated poetic quality assessment was conducted using both \texttt{GPT-4o} and \texttt{ERNIE 4.5 Turbo} as judges. Each model rated all 120 generated \textit{Songci} (covering 20 \textit{Cipai} and 6 themes) independently across fluency, coherence, and poetic quality. The final score was computed by averaging the two judge ratings. This approach provides a scalable and consistent approximation of human judgment over a large sample.

To complement the automated evaluation, we conducted a human study on a smaller, high-quality subset. For each model, we selected five \textit{Songci} with the highest combined Formal Conformity and LLM-judge scores. Each \textit{Songci} was paired with a human-written \textit{Songci} from our canonical corpus that matched the same \textit{Cipai} and theme, resulting in 30 LLM–human \textit{Songci} pairs.

Five native Chinese speakers participated in the evaluation. Each had received 19 years of Chinese-medium education, including sustained exposure to classical poetry throughout their schooling. While not literary experts, all participants demonstrated strong familiarity with the formal and aesthetic conventions of \textit{Songci}.

Results are presented in Table~\ref{table:quality}. Proprietary models \texttt{GPT-4o} and \texttt{ERNIE 4.5 Turbo} received the highest scores in all cases from both automated and human evaluations. Turing Test scores indicate that even the strongest models remain somewhat distinguishable from human poets, though the distinction is becoming increasingly subtle. Among open-source models, \texttt{DeepSeek-R1-0528-Qwen3-8B} performed best. Interestingly, while \texttt{Qwen3-32B} achieved the highest formal conformity among open-source models, it ranked among the lowest in terms of poetic quality and human judgment. This suggests that achieving strict adherence to formal constraints may come at the cost of expressive quality.

\begin{table}[h!]
\scalebox{0.7}{
\centering
\begin{tabular}{@{}l ccc@{}}
\toprule
\textbf{Model} & \textbf{Cipai ID} & \textbf{Theme Class.} & \textbf{Source Attr.} \\
& \textbf{Accuracy (\%)} & \textbf{Accuracy (\%)} & \textbf{Accuracy (\%)} \\
\midrule
\multicolumn{4}{l}{\textit{Proprietary Models}} \\
\midrule
GPT-4o & 12.61 & \textbf{75.65} & 97.67 \\
ERNIE 4.5 Turbo & 18.09 & 74.37 & \textbf{80.18} \\
\midrule
\multicolumn{4}{l}{\textit{Open-Source Models}} \\
\midrule
DeepSeek-R1-0528-Qwen3-8B & 8.41 & 51.40 & 80.95 \\
Llama-3.3-70B-Instruct & \textbf{18.80} & 49.57 & 88.64 \\
Mistral-Small-Instruct-2409 & 4.17 & 39.17 & 86.36 \\
Qwen3-32B & 10.17 & 52.54 & 86.36 \\
\bottomrule
\end{tabular}}
\caption{Results of Classification-Based Probing on each model's best-performing prompt.}
\label{tab:classification}
\end{table}

\paragraph{Classification Tasks}
To probe whether LLM-generated \textit{Songci} encode latent stylistic or thematic signals, we conducted three classification tasks using a combined dataset of 120 canonical \textit{Songci} and model-generated samples. Table~\ref{tab:classification} reports the performance of the best pipeline, which combines character-level embeddings from \texttt{bert-base-chinese} with a support vector machine (SVM) classifier.

\textbf{Cipai identification} remains the most challenging task. Accuracy across models is modest due to the fine-grained 20-class label space and small training set. The best result comes from \texttt{Llama-3.3-70B-Instruct} (18.80\%), with proprietary models like \texttt{ERNIE 4.5 Turbo} also performing relatively well (18.09\%). These results suggest that stronger models better preserve stylistic cues related to \textit{Cipai}, though the signal is still weak.

\textbf{Theme classification} shows wide variation across models. \texttt{GPT-4o} and \texttt{ERNIE 4.5 Turbo} reach 75.65\% and 74.37\% respectively, indicating strong thematic alignment in their generations. Open-source models perform substantially worse, with accuracies ranging from 39.17\% to 52.54\%. Given the balanced 6-class setup, this suggests that thematic fidelity is a key differentiator between proprietary and open-source systems.

A lower accuracy in \textbf{Source Attribution} is preferable, as it suggests the generated \textit{Songci} are more human-like. We trained a binary classifier to distinguish between model-generated and canonical \textit{Songci} using an 80/20 train-test split. Surprisingly, \texttt{GPT-4o} was the most easily identified (97.67\% accuracy), indicating that its outputs carry detectable machine-like features. In contrast, \texttt{ERNIE 4.5 Turbo} (80.18\%) and \texttt{DeepSeek-R1-0528-Qwen3-8B} (80.95\%) were harder to classify, implying a closer stylistic alignment with human-written texts.

All classification tasks use balanced label distributions (20 \textit{Cipai}, 6 themes, 2 sources). While limited by the small training set, these results provide exploratory insight into how different models encode formal and thematic signals. We interpret this analysis as complementary to the evaluation metrics presented earlier.



\section{The Generate-Critic Framework for Constrained Generation}
\begin{figure}[h]
\centering
\includegraphics[width=0.45\textwidth]{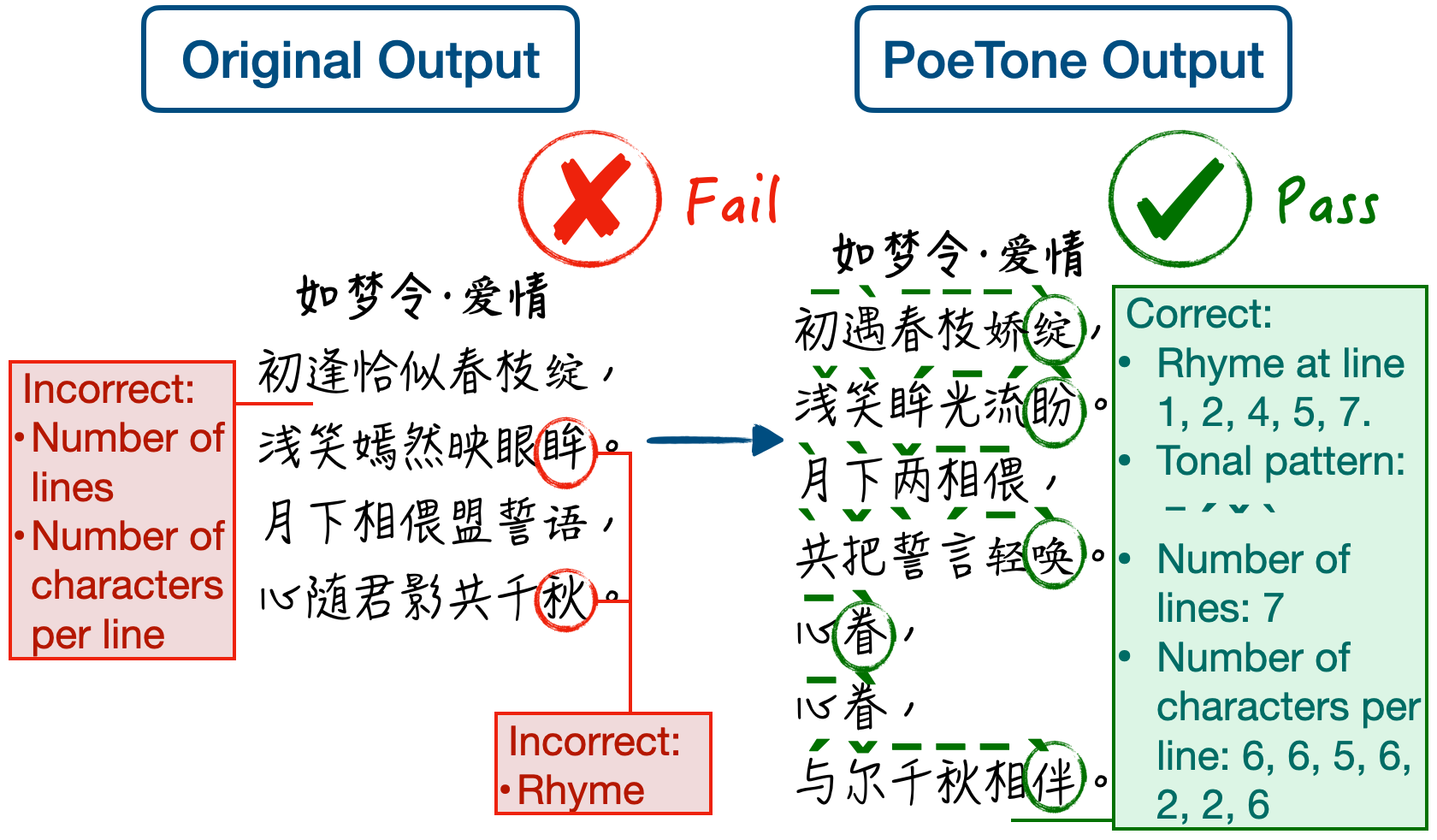}
\caption{Expected formal improvements in generated \textit{Songci} before and after fine-tuning.}
\label{fig:qualitative_example}
\end{figure}

To improve the formal conformity of generated \textit{Songci}, we adopt a Generate-Critic framework (see Figure~\ref{fig:qualitative_example} for expected improvements). This architecture consists of two core components: a Generator that produces candidate \textit{Songci}, and a Critic that evaluates their conformity against rule-based constraints.

\paragraph{The Generator} The Generator $G$ is a large language model with policy $\pi_{\theta}$, parameterized by weights $\theta$. Given an input prompt $p$ (specifying the \textit{Cipai} and theme), the Generator samples a sequence of tokens $y$ representing a complete \textit{Songci}:
\begin{equation}
    y \sim \pi_{\theta}(p)
\end{equation}

\paragraph{The Critic}
The Critic $C$ is a deterministic, rule-based function that evaluates the formal validity of a \textit{Songci} $y$ under the constraints $c$ associated with the target \textit{Cipai}. It outputs a scalar score $S$, computed as a weighted average of three sub-scores:
\begin{equation}
    S = C(y, c) = w_S \cdot S_{\text{structure}} + w_T \cdot S_{\text{tonal}} + w_R \cdot S_{\text{rhyme}}
\end{equation}
where $w_S + w_T + w_R = 1$.

\subsection{Fine-Tuning with Automated Rule-Based Feedback}
To enhance the model's ability to internalize formal constraints, we apply a rejection sampling fine-tuning strategy known as \textbf{Best-of-N Sampling (BoN)}. This approach curates a high-quality dataset using the Critic as an automated filter and then fine-tunes the Generator on these filtered examples via standard supervised learning.

The iterative process proceeds as follows:
\begin{enumerate}
    \item \textbf{Candidate Generation.} For each prompt $p_i$ in the prompt set $P$, we use the base model's policy $\pi_{\theta_{\text{base}}}$ to generate $N$ candidate \textit{Songci}:
    \begin{equation}
        Y_i = \{y_{i,1}, y_{i,2}, \dots, y_{i,N}\}, \quad \text{where } y_{i,j} \sim \pi_{\theta_{\text{base}}}(p_i)
    \end{equation}
    
    \item \textbf{Critic Scoring.} Each candidate $y_{i,j} \in Y_i$ is scored by the Critic to produce a conformity score:
    \begin{equation}
        S_{i,j} = C(y_{i,j}, c_i)
    \end{equation}

    \item \textbf{Best Sample Selection.} The highest-scoring \textit{Songci} $y_i^*$ from each set is selected:
    \begin{equation}
        y_i^* = \arg\max_{y \in Y_i} C(y, c_i)
    \end{equation}

    \item \textbf{Dataset Construction.} The selected prompt-\textit{Songci} pairs are collected into a curated dataset:
    \begin{equation}
        D_{\text{best}} = \{(p_1, y_1^*), (p_2, y_2^*), \dots, (p_M, y_M^*)\}
    \end{equation}
    where $M$ is the total number of prompts.

    \item \textbf{Supervised Fine-Tuning.} The Generator is fine-tuned on $D_{\text{best}}$ using the standard negative log-likelihood objective:
    \begin{equation}
        \mathcal{L}_{\text{SFT}}(\theta) = - \sum_{(p, y^*) \in D_{\text{best}}} \log \pi_{\theta}(y^* \mid p)
    \end{equation}

\end{enumerate}

\subsubsection{Training Details}
The method was tested on 3 models: \texttt{Qwen3-8B}, \texttt{Llama-3.1-8B-Instruct}, and \texttt{Mistral-Small-Instruct-2409}. We used the PEFT framework for Low-Rank Adaptation (LoRA), with a rank of 16 and scaling factor $\alpha = 32$. The model was quantized to 4-bit precision using \texttt{bitsandbytes} for memory-efficient training. The optimization used paged AdamW with 8-bit weights, a learning rate of $5 \times 10^{-5}$, batch size of 2, and trained for 3 epochs.

\subsection{Generate-Critic Fine-Tuned Results}
Table~\ref{tab:prompt-deltas} presents the formal conformity scores after fine-tuning using the Generate-Critic framework for 3 iterations, along with the corresponding score changes relative to the base model. All three models demonstrate measurable gains across most prompting strategies, confirming the effectiveness of rule-based automated supervision.

\texttt{Qwen3-8B} achieved the highest overall scores, both before and after fine-tuning. It exhibited particularly strong gains in zero-shot (+5.70), one-shot (+5.88), and completion (+5.63) settings. These results suggest that models with stronger baseline alignment to Chinese linguistic structure can not only maintain their advantage post-fine-tuning, but also benefit more from targeted feedback. Furthermore, high-quality initial generations provide more informative signal for the critic, such models stand to gain the most from Generate-Critic fine-tuning.

\texttt{Mistral-Small-Instruct-2409} showed notable improvements in zero-shot (+2.30) and one-shot (+3.40) settings. However, its performance declined under instruction (–1.46) and CoT (–1.13) prompts. \texttt{Llama-3.1-8B-Instruct} demonstrated consistent, moderate gains across all prompting strategies, with improvements ranging from +1.11 to +1.63. These uniform gains indicate that the Generate-Critic approach is broadly applicable, even for models not explicitly trained on Chinese or poetic tasks. The method generalizes well across prompt types, providing robust structural enhancements without degrading output diversity.


\begin{table}[ht!]
\centering
\scalebox{0.8}{
\begin{tabular}{@{}l cccc@{}}
\toprule
\textbf{} & \textbf{Llama-3.1-8B} & \textbf{Mistral-Small} & \textbf{Qwen3-8B} \\
\midrule
\textbf{Zero-shot} & 36.83 (+1.63) & 39.41 (+2.30) & 65.89 (+5.70) \\
\textbf{One-shot} & 55.53 (+1.11) & 57.81 (+3.40) & 70.41 (+5.88) \\
\textbf{Completion} & 54.77 (+1.37) & 55.31 (+1.89) & 69.38 (+5.63) \\
\textbf{Instruction} & 39.93 (+1.21) & 38.96 (–1.46) & 54.31 (+0.33) \\
\textbf{CoT} & 41.30 (+1.17) & 34.13 (–1.13) & 59.47 (–3.11) \\
\bottomrule
\end{tabular}}
\caption{Formal conformity scores after fine-tuning}
\label{tab:prompt-deltas}
\end{table}

\section{Conclusion}

This paper presents \textbf{PoeTone}, the first large-scale, multi-faceted evaluation of large language models (LLMs) in the context of constrained Chinese \textit{Songci} generation. We introduce a structured evaluation framework grounded in traditional poetic principles, benchmark 18 proprietary and open-source models across five prompting strategies, and systematically assess both formal conformity and lyrical quality.

Our results reveal that proprietary models like \texttt{GPT-4o} and \texttt{ERNIE 4.5 Turbo} outperform open-source models across most metrics, especially under zero- and one-shot prompts. However, even the best models face trade-offs between structural accuracy and poetic expressiveness.

Additionally, we propose a novel Generate-Critic framework where rule-based feedback is used to fine-tune model outputs. Applied to 3 open-source LLMs, this method yields measurable improvements in formal conformity, validating the use of automated critics for constrained generation.

Beyond \textit{Songci}, our framework has broader implications for structured text generation in domains such as legal writing, classical verse composition (e.g., sonnets, qasida, shloka), and digital humanities. By operationalizing formal constraints as reusable feedback signals, PoeTone provides a scalable, annotation-free pathway to align LLMs with symbolic, cultural, or rule-based goals, bridging human creativity and machine generation in highly structured genres.

\section*{Acknowledgments}

The authors acknowledge the financial support by the Federal Ministry of Research, Technology and Space of Germany and by Sächsische Staatsministerium für Wissenschaft, Kultur und Tourismus in the programme Center of Excellence for AI-research ``Center for Scalable Data Analytics and Artificial Intelligence Dresden/Leipzig", project identification number: ScaDS.AI.

Additionally, the project was supported by the German Federal Ministry of Research, Technology and Space (BMFTR) via the Software Campus project (01|S23070).

The authors gratefully acknowledge the computing time made available to them on the high-performance computer at the NHR Center of TU Dresden. This center is jointly supported by the Federal Ministry of Research, Technology and Space of Germany and the state governments participating in the NHR (www.nhr-verein.de/unsere-partner).

\bibliography{aaai2026}


\end{document}